\documentclass[10pt,conference]{IEEEtran}
\IEEEoverridecommandlockouts
\usepackage{cite}
\usepackage{amsmath,amssymb,amsfonts}
\usepackage{algorithmic}
\usepackage{graphicx}
\usepackage{textcomp}
\usepackage{xcolor}
\usepackage{booktabs}
\def\BibTeX{{\rm B\kern-.05em{\sc i\kern-.025em b}\kern-.08em
    T\kern-.1667em\lower.7ex\hbox{E}\kern-.125emX}}
\begin{document}


\title{URA-NER: A Unified Retrieval-Augmented Framework with Retrieval Alignment and Uncertainty Reduction for Low-Resource NER}

\author{\IEEEauthorblockN{Jingyu Wang$^{1}$, Shijie Wu$^{1}$, Fusheng Jin$^{1, *}$}
\IEEEauthorblockA{$^{1}$School of Computer Science and Technology, Beijing Institute of Technology, Beijing, China\\
$^{*}$Corresponding author\\
jyw.bit.2003@gmail.com, \{wsj1595, jfs21cn\}@bit.edu.cn}}
\maketitle

\begin{abstract}
In-context learning (ICL) based on large language models (LLMs) has shown promising potential in alleviating performance bottlenecks caused by the limited availability of annotated data in Named Entity Recognition (NER). 
However, existing methods still face issues of retrieval misalignment and generation uncertainty, making their performance heavily dependent on the LLM's capabilities.
As the parameter scale of LLMs decreases, their performance in few-shot settings deteriorates significantly. In this paper, we propose a novel unified retrieval-augmented framework, URA-NER, including three key components: Progressive Granularity Retrieval (PGR), Model-aware Representation Enhancement (MaRE), and Reason-aware Knowledge Verification. PGR is a two-stage retrieval mechanism that achieves stage alignment. It first retrieves demonstrations for span detection based on the query's global semantics, and then for type classification based on the specific entity context, providing fine-grained local information. Moreover, MaRE employs entity pre-recognition to guide the construction of representations, ensuring the query and demonstrations are aligned within the LLM’s semantic space and attention pattern. In addition, to mitigate generation uncertainty, we propose RaKV, a closed-loop ``generation-retrieval-verification'' process. It explicates the LLM's reasoning paths, leverages them for the retrieval of external knowledge, and reorganizes the knowledge into verification evidence aligned with the original reasoning paths. We conduct extensive experiments on multiple low-resource NER datasets. Results demonstrate that URA-NER significantly enhances the performance of LLMs under low-resource settings, with particularly pronounced gains for smaller LLMs, achieving new state-of-the-art results on several benchmarks.
\end{abstract}


\section{Introduction}
Named Entity Recognition (NER) is a fundamental task in information extraction, aiming to identify entities of predefined types from unstructured text~\cite{1, 2}, such as persons, locations, and organizations. As a crucial step in text understanding, NER plays a vital role in a wide range of natural language processing (NLP) tasks, including knowledge graph construction~\cite{44, 45, 39}, question answering~\cite{18, 19, 20}, and information retrieval~\cite{21, 22}. With sufficient annotated data, existing neural network-based models have achieved remarkable progress on NER through supervised learning approaches~\cite{3, 4, 5, 16}. However, in complex scenarios where annotated data is scarce, models struggle to acquire enough data to learn effective entity recognition capabilities, leading to a substantial drop in performance.

Recently, researchers have begun to explore the use of large language models (LLMs) \cite{6,7,8,9,13} for NER, transforming the sequence labeling task into a generation task. Some studies attempt to guide LLMs to perform NER via in-context learning (ICL), where a small number of annotated demonstrations are incorporated into the input as context to help the LLM learn the task effectively. This approach shows promising performance under low-resource settings. Early studies \cite{6,46} adopt a fixed context. They provide the same demonstrations for different query texts and achieve notable improvements.
However, this approach lacks dynamic interaction with the query text, making it difficult to capture semantic alignment across different inputs and thus limiting the model’s performance. To address this issue, some studies \cite{12,13,14,15} have introduced retrieval-augmented and dynamic demonstrations. They build a pool of annotated demonstrations and retrieve the most relevant ones for each query text using retrieval algorithms.
The retrieved demonstrations help LLMs identify similar entity boundaries and types in the query text. Current methods depend on the intrinsic abilities of LLMs. Although large-scale LLMs (eg, GPT, Gemini) achieve strong performance, smaller models exhibit notable degradation in low-resource scenarios.

Current methods still face several key challenges that limit the performance of LLMs on low-resource NER:

(1) \textbf{Lack of stage alignment in the retrieval mechanism.}
Existing retrieval methods typically use a single granularity, such as retrieving based on the entire text or specific segments within the text, which fails to meet the stage-specific requirements of NER. Span extraction requires global context to identify all potential entities in the text, while type classification relies on local context around each entity. This mismatch leads to a misalignment between retrieved demonstrations and the objectives of each stage.

(2) \textbf{Lack of representation alignment with the LLM.}
Existing retrieval methods rely on independent encoders (e.g., BERT, BGE) to generate text representations. However, these encoders differ from the LLM performing NER in terms of pretraining objectives and attention mechanisms, leading to a misalignment in representation spaces. As a result, the retrieved demonstrations may deviate from the LLM's semantic understanding, introducing potential bias.

(3) \textbf{Generation uncertainty of LLMs.}
In low-resource scenarios, LLMs with relatively small parameter scales often generate unstable outputs due to limited internal knowledge or biased in-context learning. The same query text may lead to inconsistent results across different generations. Existing methods lack effective validation mechanisms during the generation process, making it difficult to identify and correct reasoning errors or deviations in model predictions.

To address the challenges, we propose a novel unified retrieval-augmented framework, \textbf{URA-NER}, which focuses on retrieval alignment and uncertainty Reduction in LLM generation.
(1) We introduce \textbf{Progressive Granularity Retrieval (PGR)} to achieve stage alignment through a two-stage ``global-to-local'' retrieval strategy. In the first stage, global demonstrations with similar entity distribution are retrieved based on the global semantics of the query text, providing macro-level guidance for span extraction. In the second stage, for each extracted entity span, demonstrations with more accurate type annotations are retrieved by focusing on the entity’s local context, offering micro-level support for type classification.
(2) Furthermore, we incorporate \textbf{Model-aware Representation Enhancement (MaRE)} into the retrieval process to align the LLM’s semantic space and attention pattern. Specifically, we construct text representations that reflect the LLM's inherent semantic preferences through entity pre-recognition by leveraging its autoregressive attention and internal representation.
(3) Lastly, we propose \textbf{Reason-aware Knowledge Validation (RaKV)} to mitigate generation uncertainty through a ``generation–retrieval–validation'' process. The LLM is instructed to generate reasoning paths. External knowledge is then retrieved based on key semantic nodes in the reasoning paths and reorganized following the paths, forming evidence that closely aligns with the LLM’s internal logic and inference process to facilitate self-validation of the outputs.

We conduct extensive experiments on the five low-resource subsets (Science, Politics, Music, Literature, and AI) of the CrossNER dataset~\cite{40} and general-domain CoNLL2003 dataset~\cite{41}. Results show that URA-NER achieves new state-of-the-art performance across all five CrossNER subsets under low-resource settings, outperforming even methods that rely on large-scale LLMs like GPT, despite running on smaller LLMs.
This significantly reduces dependence on the LLM scale, demonstrating the effectiveness and generalization of URA-NER in low-resource NER.

The code and data for this work have be made publicly available: https://github.com/2025-NLP/NER.

\section{Related Work}
\subsection{Named Entity Recognition}
With the development of deep learning, neural network-based methods have become mainstream. Models such as LSTM~\cite{24} and those ~\cite{26, 27, 28} based on the transformer architecture~\cite{25} can automatically learn contextual semantic information and have achieved significant performance. In high-resource settings, neural networks~\cite{29,30} have shown remarkable effectiveness in NER. However, under low-resource scenarios where annotated training data is scarce, these models often suffer substantial performance degradation. Large Language Models (LLMs), endowed with rich prior knowledge and strong language understanding capabilities, have demonstrated superior performance in low-resource NER. Recent studies based on LLMs can be broadly divided into two categories: one uses LLMs as auxiliary tools for smaller models, enabling data augmentation~\cite{7, 31, 32}, knowledge distillation~\cite{33, 34}, etc.; the other directly employs LLMs as the backbone through instruction tuning~\cite{8, 14} or in-context learning~\cite{6, 12, 13}. In this paper, we focus on the latter.

\subsection{In-Context Learning}
Brown et al.~\cite{35} propose in-context learning (ICL), a method that enables LLMs to perform tasks without updating parameters by providing a few annotated demonstrations in the input. Due to its strong few-shot generalization ability, ICL has been widely applied in various NLP tasks~\cite{36,37,38}. In NER, recent efforts focus on optimizing the demonstration retrieval strategy, on how to retrieve the most beneficial demonstrations from the training set to improve performance. Shiraishi et al.~\cite{12} adopt BM25 and [CLS] token-based global semantic representations from BGE to measure relevance. Wang et al.~\cite{13} and Nandi et al.~\cite{14} use token-level and word-level granularity combined with the KNN strategy for finer-grained semantic matching. Wu et al.~\cite{15} further propose a dual relevance measure combining ontology and contextual consistency. These methods improve the performance of LLMs on low-resource NER to varying degrees, highlighting the importance of retrieving beneficial demonstrations.

\subsection{Differences between URA-NER and Previous Methods}
In this paper, we focus on the optimization of the retrieval mechanism. Unlike previous studies that adopt single-stage retrieval, we propose PGR that aligns with the distinct characteristics of the two phases in NER through a two-stage design, thus achieving stage alignment. Moreover, unlike prior methods that rely solely on independent encoders for text representations, we introduce MaRE, which constructs representation vectors using the LLM’s attention patterns to achieve representation alignment. Finally, we propose RaKV, which retrieves external knowledge using reasoning paths generated by the LLM to validate predictions and reduce generation uncertainty. In contrast to prior work that relies on either closed-source large LLMs or fine-tuned small LLMs, URA-NER achieves substantial improvements on small LLMs without fine-tuning, reducing dependence on model scale.

\section{Methodology}

\begin{figure*}
    \centering
    \includegraphics[width=1\linewidth]{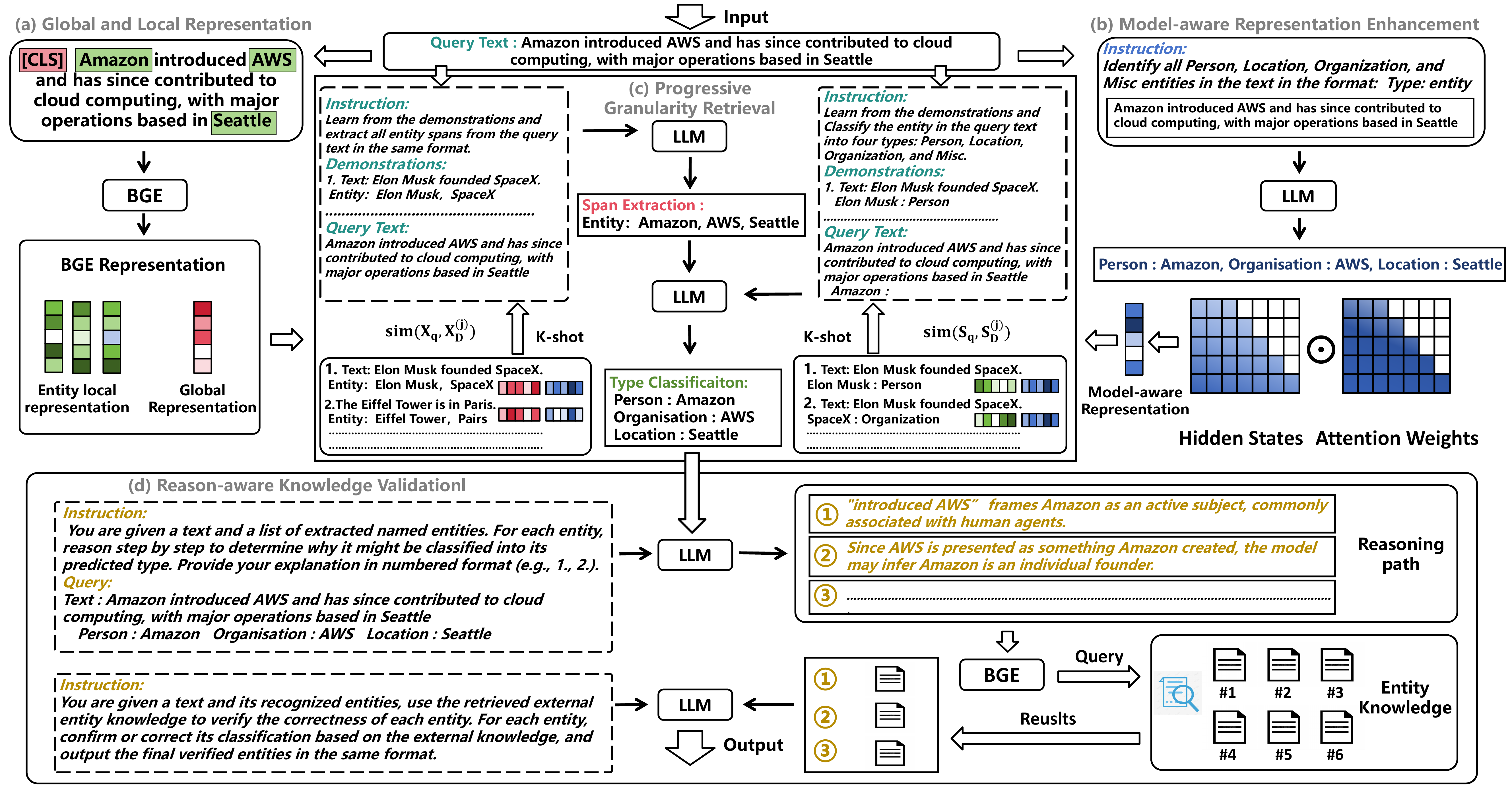}
    \caption{Architecture overview of URA-NER.}
    \label{Framework}
\end{figure*}

\subsection{Task Description}
Given a query text \( X_q = \{x^1_q, x^2_q, \ldots, x^n_q\} \), which represents a text composed of $n$ tokens, the objective of NER is to identify all named entities in the text, formally represented as an entity set \( E_q = \{(s^{(i)}_q:y^{(i)}_q)\}_{i=1}^N \), where \( s^{(i)}_q \) denotes the span of the $i$-th entity and \( y^{(i)}_q \) represents its type belonging to a predefined type set $Y$. In in-context learning, multiple annotated demonstrations are concatenated with the query text to construct contextual prompts that guide the LLM in performing NER. For methods employing dynamic retrieval to construct contextual prompts, there exists a candidate demonstration pool \( D = \{(X_d^{(i)}, E_d^{(i)})\}_{i=1}^M \), where each \( X_d^{(i)} \) represents an demonstration text and \( E_d^{(i)} \) denotes its entity annotations. Based on a specific retrieval mechanism \( \Phi(\cdot) \), the $k$ most relevant demonstrations to the current query text \( X_q \) are retrieved from $D$, forming a demonstration subset \( D' \subseteq D \). Finally, the subset $D'$ is concatenated with the query text \( X_q \) to construct contextual prompts that guide the LLM.

\subsection{Architecture Overview}
Figure~\ref{Framework} illustrates the overall framework of \texttt{URA-NER}, which processes a given query text $X_q$ through four steps:  
(a) We obtain both global and local representations of $X_q$ using the BGE encoder;  
(b) Model-aware representations are constructed via the MaRE module;  
(c) A two-stage reasoning process is conducted by PGR, where entity span extraction and type classification are performed sequentially to obtain initial NER predictions; 
(d) Finally, the PaKV verifies the predicted entities against external knowledge, obtaining the final NER output.

\subsection{Model-aware Representation Enhancement}
Prior studies~\cite{12,14,15} rely solely on independent encoders (e.g., BERT, BGE) to compute textual semantic representations. which decouples from the actual LLM performing NER, where different LLMs possess different attention patterns and representation spaces. To ensure representation alignment in the retrieval mechanism, we propose MaRE.

\subsubsection{Pre-recognition}
LLMs often rely on specific key tokens when recognizing entities in text, which reflect their inherent attention patterns. Inspired by research~\cite{15}, we introduce entity pre-recognition. By guiding the LLM to identify potential entities autoregressively under the zero-shot setting, we progressively extract the model's attention distribution across different tokens. For the query text \( X_q \), we obtain pre-recognition results \( Y_T \) after \( T \) time steps:
\begin{equation}
Y_T = F_{LLM}(X_q, I_0),
\end{equation}
where \( F_{LLM}(\cdot) \) represents the LLM model, and \( I_0 \) is the instruction for pre-recognition, as shown in Figure~\ref{Framework}.

\subsubsection{Model-aware Representation Construction}
We construct model-aware representations based on the LLM's pre-recognition process. The LLM's input consists of the query text and instruction \([X_q, I_0] = \{x^1_q, \ldots, x^n_q,  x^j_q, \ldots, x^m_j\}\). At timestep t, the sequence already generated by the LLM is \(Y_{t \leq T} = \{y^1, y^2, \ldots, y^{t-1}\}\). For decoder-only architecture LLMs, we treat the original input \([X_q, I_0]\) and the generated sequence \(Y_t\) as an integrated input \([X_q, J_0, Y_{t \leq T}]\). Through encoding, we extract the final hidden layer vectors \(\mathcal{H}
_{llm}^{(t)} \in \mathbb{R}^{(n+m+t-1) \times d_0}\) of the query text $X_q$:
\begin{equation}
\mathcal{H}
_{llm}^{(t)} = \mathcal{F}_{Embed}([X_q, I_0, Y_{t \leq T}]),
\end{equation}
where \(d_0\) is the dimension of the LLM's final hidden layer, and \(\mathcal{F}_{Embed}(\cdot)\) represents the model's hidden vector computation function. In this way, we obtain semantic embeddings in the LLM's representation space.

To further analyze which tokens the LLM focuses on at timestep t, we extract the multi-head attention weights from the model's final layer \(A_t = \{a_t^{(i)}\}_{i=1}^{p}\):
\begin{equation}
a_t^{(i)} = \mathcal{F}_{Attention}(\mathcal{H}
_{llm}^{(t)}),
\end{equation}
where $a_t^{(i)} \in \mathbb{R}^{1 \times (n+m+t-1)}$ is the attention weights of the $i$-th head at timestep $t$, and $\mathcal{F}_{Attention}(\cdot)$ denotes the LLM's attention weight computation function.

Different attention heads focus on distinct sequence parts. We balance them through mean pooling of attention weights. Tokens with high attention weights carry more critical semantic information. We compute weighted sums of hidden vectors $\mathcal{H}
_{llm}^{(t)}$ using attention weights $\bar{a}_t$ as model-aware representation at each timestep $t$, obtaining $ \hat{\mathcal{H}
}{_{llm}} = \{\hat{h}_{llm}^{(t)}\}_{t=1}^T$:
\begin{equation}
\hat{h}_{llm}^{(t)} = \left( \frac{1}{|A_t|} \sum_{a_t^{(i)} \in A_t} a_t^{(i)} \right) \cdot \mathcal{H}
_{llm}^{(t)},
\end{equation}
where $\hat{h}_{llm}^{(t)} \in \mathbb{R}^{1 \times d_0}$ is the model-aware representation of the token generated at timestep $t$. To mitigate the influence of stopwords in the LLM's generated output, we remove these tokens' model-aware representations to obtain $\hat{\mathcal{H}
}^{\prime}_{llm}$. We apply mean pooling to the token representations in $\hat{\mathcal{H}
}^{\prime}_{llm}$:

\begin{equation}
{h}^{q}_{llm} = \frac{1}{|\hat{\mathcal{H}
}^{\prime}_{llm}|} \sum_{\hat{h}_{llm} \in \hat{\mathcal{H}
}^{\prime}_{llm}} \hat{h}_{llm}.
\end{equation}

Through this process, we obtain the model-aware representation ${h}^{q}_{llm}$ for $X_q$. For each demonstration text in $\mathcal{D}$, we apply the same procedure to obtain its model-aware representations, which are precomputed and stored locally to support efficient retrieval and ensure representation alignment.

\subsection{Progressive Granularity Retrieval}
Although we build model-aware representations via the LLM’s pre-recognition for consistency, the LLM mainly focuses on contextual generation, not explicit semantic alignment. Therefore, we also use the BGE model to compute semantic embeddings. We insert a \texttt{[CLS]} token at the start of \(X_q\) and input it into BGE, extracting the final hidden layer vectors
\(\mathcal{H}
_{bge}^q=\left[ h_{[cls]}^q, h_{1}^q, \ldots, h_{n}^q \right] \in \mathbb{R}^{(n+1) \times d_1}\):
\begin{equation}
\mathcal{H}
_{{bge}}^q = \mathcal{F}_{BGE}([CLS], X_q),
\end{equation}
where $h_{[cls]}^q$ is the global semantic features of the query text, $\mathcal{F}_{BGE}(\cdot)$ denotes the BGE model, and $d_1$ is the dimension of BGE's final hidden layer.

\subsubsection{Span Extraction}
In span extraction, LLMs focus more on the global semantics of the query text $X_q$. We compute BGE's global semantic similarity and model-aware representation similarity between $X_q$ and each demonstration in $D$ using cosine similarity, with different weights assigned:

\begin{equation}
\begin{split}
sim(X_q, X_D^{(j)}) &= \lambda_0 \cdot \cos({h}^{q}_{llm}, {h}^{(j)}_{llm})\\
&\quad + (1 - \lambda_0) \cdot \cos\left(h^q_{[cls]}, h^{(j)}_{[cls]}\right),
\end{split}
\end{equation}
where $\cos$ denotes the cosine similarity function and $\lambda_0$ is the weight for model-aware representation similarity in the span extraction phase, the similarity score $sim$ measures the relevance between the query and each demonstration. We retrieve the top-$k$ most relevant demonstrations as context, based on which the LLM generates a set $S_q$ containing $N^{\prime}$ potential entity spans $S_q$.

\subsubsection{Type Classification}

Unlike span extraction, the type classification phase focuses more on entities themselves and their contextual information. For each entity span \(s_q \in S_q\), we extract its token vectors from $\mathcal{H}
_s \in \mathcal{H}
_{{bge}}^q$ and average them to obtain the semantic representation $h_{span}$ of the entity span:
\begin{equation}
h_{span} = \frac{1}{|\mathcal{H}
_s|} \sum_{h \in \mathcal{H}
_s} h.
\end{equation}

For \(\mathcal{D}\), where each demonstration has annotated entity spans, we compute semantic representations for all entities in the same way. We use cosine similarity to compute both the entity similarity and model-aware representation similarity between entity spans in the query text and each entity in the example library, with different weights assigned:
\begin{equation}
\begin{split}
sim\left( s_q, s_\mathcal{D}^{(j)} \right) &= \lambda_1 \cdot \cos({h}^{q}_{llm}, {h}^{(j)}_{llm})\\
&\quad + (1 - \lambda_1) \cdot \cos(h_s, h_{s}^{(j)}),
\end{split}
\end{equation}
where $\lambda_1$ is the weight for model-aware representation similarity in the type classification phase. We select the top-$k$ most relevant demonstrations for the LLM to classify each entity span. By classifying each entity span $s_q$ in the query text $X_q$, the LLM outputs the final NER result $E_q = \left\{ \left( s_q^{(i)}, y_q^{(i)} \right) \right\}_{i=1}^{N^{\prime}}$.


\subsection{Reason-aware Knowledge Validation}

LLMs exhibit inherent generation uncertainty, particularly in low-resource scenarios. The lack of sufficient prior knowledge and contextual demonstrations may cause LLMs to generate inconsistent or even contradictory outputs. To address this, we propose the RaKV.

\subsubsection{Reasoning Path Explicitness}

The reasoning path captures the LLM's decision process during task execution, revealing its focus and the basis for its predictions. By making these paths explicit, we can use them as cues to retrieve relevant information from external knowledge bases for verification.  
Given a query text \(X_q\) and its recognition results \(E_q = \left\{ (s_q^{(i)}, y_q^{(i)}) \right\}_{i=1}^{N^{\prime}}\), we feed them into the LLM along with instruction \(I_1\) to generate reasoning paths for each entity, resulting in a set of reasoning paths $R_q=\{r^{(i)}\}_{i=1}^{N^\prime}$:
\begin{equation}
R_q = F_{LLM} \left( X_q, \left\{ (s_q^{(i)}, y_q^{(i)}) \right\}_{i=1}^{N^{\prime}}, I_1 \right).
\end{equation}

Each entity's reasoning path contains several reasoning steps, as shown in Figure~\ref{Framework}.

\subsubsection{Entity Knowledge Retrieval}

For each entity $(s_q^{(i)}, y_q^{(i)})$ in $E_q$, we retrieve relevant external knowledge from Wikipedia and segment it into knowledge chunks to ensure semantic completeness and contextual coherence. We then extract [CLS] global semantics using the BGE model for each reasoning step of  $r^{(i)}$ and knowledge chunk, selecting the most relevant knowledge chunk for each reasoning step of $r^{(i)}$ via cosine similarity. These are concatenated according to the reasoning path to form external entity knowledge $K^{(i)}$.

\subsubsection{Validation}

For each entity \((s_q^{(i)}, y_q^{(i)})\), we provide corresponding external knowledge \(K^{(i)}\) to the LLM, along with instruction \(I_2\) (as shown in Figure~\ref{Framework}), to perform verification. The LLM then outputs the validated result:
\begin{equation}
(\hat{s}_q^{(i)}, \hat{y}_q^{(i)}) = F_{LLM} \left( X_q, (s_q^{(i)}, y_q^{(i)}) , K^{(i)}, I_2 \right).
\end{equation}

Finally, we obtain the named entity recognition results $\hat{E}_q=\{\hat{s}_q^{(i)}, \hat{y}_q^{(i)}\}_{i=1}^{N^\prime}$for the query text $X_q$.

\section{Experiments}

\subsection{Datasets}
Table~\ref{tab:1} presents the statistics of the experimental datasets. We conduct experiments on a wide range of datasets, including five domain-specific subsets from the \texttt{CrossNER} dataset~\cite{40} (i.e., \texttt{Science}, \texttt{Politics}, \texttt{Music}, \texttt{Literature}, and \texttt{AI}) as well as one general-domain dataset: \texttt{CoNLL2003}~\cite{41}. Each CrossNER subset contains entity types specific to its corresponding domain, with training set sizes of 100 or 200 samples, making them suitable for evaluating low-resource NER. CoNLL2003, as a general-domain dataset, includes common entity types such as \textit{person} and \textit{location}. Given the significantly smaller training sets in CrossNER, low-resource NER becomes more challenging. Thus, we primarily focus our experiments and analysis on the five subsets of the CrossNER.

\begin{table}
\centering
\caption{The statistics of all datasets used in experiments.}
\begin{tabular}{cccccc}
\toprule
Dataset & Type Num & Train & Dev & Test \\
\midrule
Politics & 9 & 200 & 541 & 651 \\
Science & 17 & 200 & 450 & 543 \\
Music & 13 & 100 & 380 & 456 \\
Literature & 12 & 100 & 400 & 416 \\
AI & 14 & 100 & 350 & 431 \\
CoNLL2003 & 4 & 14987 & 3466 & 3684 \\
\bottomrule
\end{tabular}
\label{tab:1}
\end{table}

\subsection{Settings}
We conduct our experiments on two NVIDIA RTX A6000 GPUs with 48 GB of memory. The backbone language models are instruction-tuned versions of \texttt{Qwen2.5-7B} and \texttt{Qwen2.5-14B}~\cite{42}, which enhance the models' ability to follow input instructions. To verify the generalizability of our method, we also evaluate it on \texttt{Deepseek-R1 (API)}~\cite{43}. External entity knowledge is retrieved via the \texttt{Wikipedia API}. Both the semantic representations of the demonstration pool and the external entity knowledge are stored locally using \texttt{FAISS-GPU} to improve retrieval efficiency. Following the SOTA baseline~\cite{12}, we use the same pre-trained \texttt{bge-base-en} model as our external encoder. 

Following previous studies~\cite{12,13,15}, we construct the demonstration pool from all samples in the training set. Based on ablation studies, we set the hyperparameters $\lambda_1$ to 0.5 and $\lambda_2$ to 0.3, respectively, which achieve the best performance. We conduct experiments under both 1-shot and 5-shot settings. We adopt micro-F1 as the evaluation metric and report the average and standard deviation over five runs on the test set.

\begin{table*}
\centering
\caption{Micro-F1 of different methods under various k-shot settings. A smaller k-shot means fewer demonstrations. Bold: best results.}
\begin{tabular}{lc*{6}{c}}
\toprule
Method & k-shot & Science & Politics & Music & Literature & AI & CoNLL2003 \\
\midrule
PromptNER (GPT3.5) & 2 & 64.83 & 71.74 & 77.78 & 64.15 & 59.35 & 78.62 \\
GPT-NER (GPT-3) & 1-32 & 70.77 & 74.71 & 78.30 & 62.18 & 66.07 & \textbf{90.91} \\
PromptNER (GPT4) & 2 & 72.59 & 78.61 & 84.26 & 74.44 & 64.83 & 83.48 \\
ConsistNER (GPT-3.5) & 5 & -- & -- & -- & -- & -- & 78.87 \\
IF-WRANER (LLaMA) & 5 & 75.31 & 79.80 & 85.43 & 75.52 & 68.81 & -- \\
RENER (Gemini-1.5-pro) & 13-21 & 78.43 & 82.02 & 83.21 & 72.11 & 72.09 & 86.10 \\
Deepseek-R1 + [CLS] & 10 & 77.89±0.3 & 82.52±0.1 & 85.06±0.2 & 76.91±0.2 & 72.58±0.3 & 86.79±0.2 \\
\midrule
Qwen2.5-7B + URA-NER & 1 & 69.58±0.5 & 73.64±0.4 & 79.96±0.3 & 67.29±0.4 & 61.67±0.4 & 77.23±0.3 \\
Qwen2.5-14B + URA-NER & 1 & 73.38±0.4 & 77.21±0.2 & 82.37±0.3 & 71.86±0.5 & 66.50±0.3 & 81.62±0.2 \\
Qwen2.5-7B + URA-NER & 5 & 77.76±0.3 & 80.40±0.3 & 83.43±0.4 & 74.56±0.4 & 72.82±0.3 & 85.59±0.2 \\
Qwen2.5-14B + URA-NER & 5 & \textbf{79.41±0.3} & \textbf{83.97±0.2} & \textbf{86.02±0.2} & \textbf{77.23±0.3} & \textbf{74.55±0.2} & 87.02±0.2 \\
\bottomrule
\end{tabular}
\label{tab:2}
\end{table*}

\begin{table*}
\centering
\caption{Ablation study of Qwen2.5-7B under the 5-shot setting. Bold: best results.}
\begin{tabular}{l*{6}{c}}
\toprule
Method & Science & Politics & Music & Literature & AI & CoNLL2003 \\
\midrule
Qwen2.5-7B + [CLS] & 68.45$\pm$0.6 & 69.92$\pm$0.6 & 75.41$\pm$0.4 & 64.77$\pm$0.7 & 58.73$\pm$0.8 & 79.83$\pm$0.4 \\
Qwen2.5-7B + token kNN & 69.31$\pm$0.6 & 70.48$\pm$0.7 & 75.65$\pm$0.5 & 67.52$\pm$0.6 & 60.97$\pm$0.7 & 80.42$\pm$0.3 \\
\midrule
Qwen2.5-7B + PGR & 71.18$\pm$0.7 & 73.57$\pm$0.6 & 77.51$\pm$0.5 & 69.09$\pm$0.7 & 63.54$\pm$0.8 & 81.36$\pm$0.4 \\
Qwen2.5-7B + PGR \& MaRE (w/o LLM attention) & 71.92$\pm$0.5 & 74.65$\pm$0.7 & 77.98$\pm$0.4 & 69.27$\pm$0.5 & 64.82$\pm$0.7 & 81.94$\pm$0.3 \\
Qwen2.5-7B + PGR \& MaRE (w/ LLM attention) & 73.23$\pm$0.6 & 75.42$\pm$0.6 & 78.36$\pm$0.4 & 70.29$\pm$0.6 & 65.87$\pm$0.7 & 82.46$\pm$0.3 \\
Qwen2.5-7B + URA-NER & \textbf{77.76$\pm$0.3} & \textbf{80.40$\pm$0.3} & \textbf{83.43$\pm$0.4} & \textbf{74.56$\pm$0.4} & \textbf{72.82$\pm$0.3} & \textbf{85.59$\pm$0.2} \\
\bottomrule
\end{tabular}
\label{tab:3}
\end{table*}

\subsection{Baselines}
We select representative LLM-based ICL methods for low-resource NER as baselines:  
(1) \textbf {PromptNER}~\cite{6}: It employs statically constructed demonstration prompts and guides the GPT to perform NER via chain-of-thought reasoning.  
(2) \textbf {GPT-NER}~\cite{13}: Based on GPT-3, it performs token-level kNN matching to retrieve relevant texts containing similar tokens as demonstrations. 
(3) \textbf{ConsistNER}~\cite{15}: Introduces ontological and contextual consistency to measure textual similarity for retrieval.  
(4) \textbf {IF-WRANER}~\cite{14}: Utilizes word-level kNN matching and applies instruction tuning to the LLaMA. 
(5) \textbf{RENER}~\cite{12}: Retrieves demonstrations by computing inter-text similarity using [CLS] representations extracted via the encoder or by employing BM25. This method represents the current SOTA on the CrossNER dataset.
(6) \textbf{Deepseek-R1}~\cite{43}: We evaluate the [CLS] method from RENER~\cite{12} using the LLM Deepseek-R1 under the 10-shot setting.

\subsection{Experimental Results}
\subsubsection{Main Results}
In Table~\ref{tab:2}, we present the main results. On the five CrossNER subsets, Qwen2.5-14B achieves new SOTA results, surpassing prior methods based on larger LLMs such as GPT and Gemini. A smaller k-shot corresponds to fewer demonstrations, which increases task difficulty. Specifically, under the 5-shot setting, our method using Qwen2.5-14B outperforms RENER and Deepseek-R1 even when they use more than 10 demonstrations. In the 1-shot setting, Qwen2.5-7B and Qwen2.5-14B achieve comparable performance to PromptNER using GPT-3.5 and GPT-4 under 2-shot settings, demonstrating the robustness of our method in low-resource NER. Moreover, our 5-shot results with Qwen2.5-7B are comparable to IF-WRANER, which requires instruction tuning on LLaMA-7B and extra computation.
On the general-domain CoNLL2003 dataset, our method achieves 87.02\% F1 with Qwen2.5-14B under the 5-shot setting, slightly below GPT-NER's 90.91\% with GPT-3. We attribute this to GPT-NER’s token-level kNN retrieval strategy and the properties of CoNLL2003: its shorter texts and fewer entities allow more demonstrations to be recalled under the same $ k$-shot setting, thereby improving performance.

\subsubsection{Ablation Study}
In Table~\ref{tab:3}, we conduct an ablation study using Qwen2.5-7B to validate the contribution of each component in URA-NER. Additionally, we report the performance of two baseline methods implemented with Qwen2.5-7B: \texttt{[CLS]}~\cite{12} and \texttt{token kNN}~\cite{13}. We observe that incorporating the PGR leads to a significant performance gain over both baselines. This demonstrates that our two-stage retrieval strategy better aligns demonstrations with the distinct needs of the two NER stages. Building upon PGR, we implement MaRE in two settings - \texttt{w/ and w/o LLM attention}, controlling whether the LLM's attention weights are used in representation construction. Both settings yield additional performance gains over PGR alone. This demonstrates that considering the LLM's internal semantic space when computing demonstration relevance can reduce bias of the external encoder during retrieval. Incorporating LLM attention weights improves performance by better aligning representations with the LLM's internal attention pattern.

With the addition of RaKV, performance improves significantly. This demonstrates that RaKV effectively leverages external knowledge to verify the model’s predictions, thereby increasing accuracy. Moreover, the standard deviation decreases, indicating that RaKV can reduce generation uncertainty in low-resource NER. In addition, compared to \texttt{CoNLL2003}, the five subsets of \texttt{CrossNER} show more substantial improvements in F1. Since \texttt{CrossNER} contains longer texts and more entities, it presents a more complex context. This result suggests that in more complex contexts with higher entity density, RaKV is more important for enhancing model robustness and output reliability.

\begin{figure}
    \centering
    \includegraphics[width=0.9\linewidth]{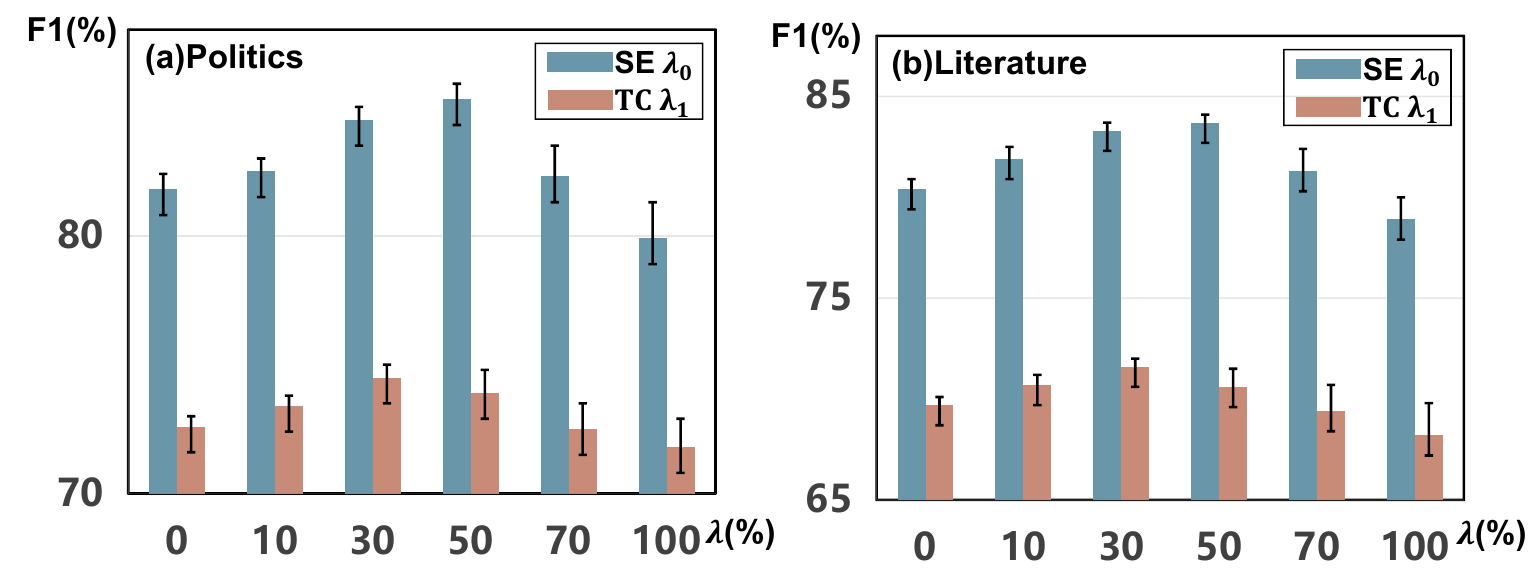}
    \caption{The performance of the Qwen2.5-7b across different $\lambda$ weights.}
    \label{fig:2}
\end{figure}

\begin{table*}
\centering
\caption{Performance of Deepseek-R1 under different configurations. Values in parentheses are gains over \texttt{+[CLS]}.}
\begin{tabular}{lcccccc}
\toprule
Method & Science & Politics & Music & Literature & AI & CoNLL2003 \\
\midrule
Deepseek-R1+URA-NER($\lambda=0$) & 78.81 (+4.46) & 81.67 (+3.56) & 84.06 (+2.00) & 75.91 (+3.23) & 73.99 (+4.92) & 85.19 (+1.90) \\
Deepseek-R1+PGR & 75.02 (+1.33) & 79.23 (+1.12) & 82.97 (+0.91) & 73.09 (+0.41) & 70.50 (+1.43) & 84.11 (+0.82) \\
Deepseek-R1+[CLS] & 74.35 & 78.11 & 82.06 & 72.68 & 69.07 & 83.29 \\
\bottomrule
\end{tabular}
\label{tab:5}
\end{table*}

\subsubsection{$\lambda$ Weights Analysis}

In Figure~\ref{fig:2}, we report the performance of Qwen2.5-7B on the dev sets of \texttt{Politics} and \texttt{Literature} under \texttt{Qwen2.5-7B + PGR\&MaRE}. We investigate the effect of different $\lambda$ weights for model-aware representational similarity on Span Extraction (SE) and Type Classification (TC). For SE, we report the F1 score of entity spans; for TC, we report the overall F1 score. As the weight increases, both F1 scores improve significantly, indicating that incorporating the LLM’s own representation space and attention patterns into the retrieval mechanism enhances the relevance of retrieved demonstrations. Notably, TC achieves its best performance when the weight is 30\%, while SE peaks at 50\%. We speculate that this is because SE relies more on the global semantics of the input, and model-aware representations enhance it. However, as the weight increases, both SE and TC experience a sharp drop in F1 scores and a rise in standard deviation. The generation of LLMs under the zero-shot setting is unstable, with outputs varying across multiple runs. This variability leads to fluctuations in the computed model-aware representations. When $\lambda$ is high, the instability is further amplified, indicating the necessity to incorporate the fixed representation from the external encoder.
\begin{figure}
    \centering
    \includegraphics[width=0.9\linewidth]{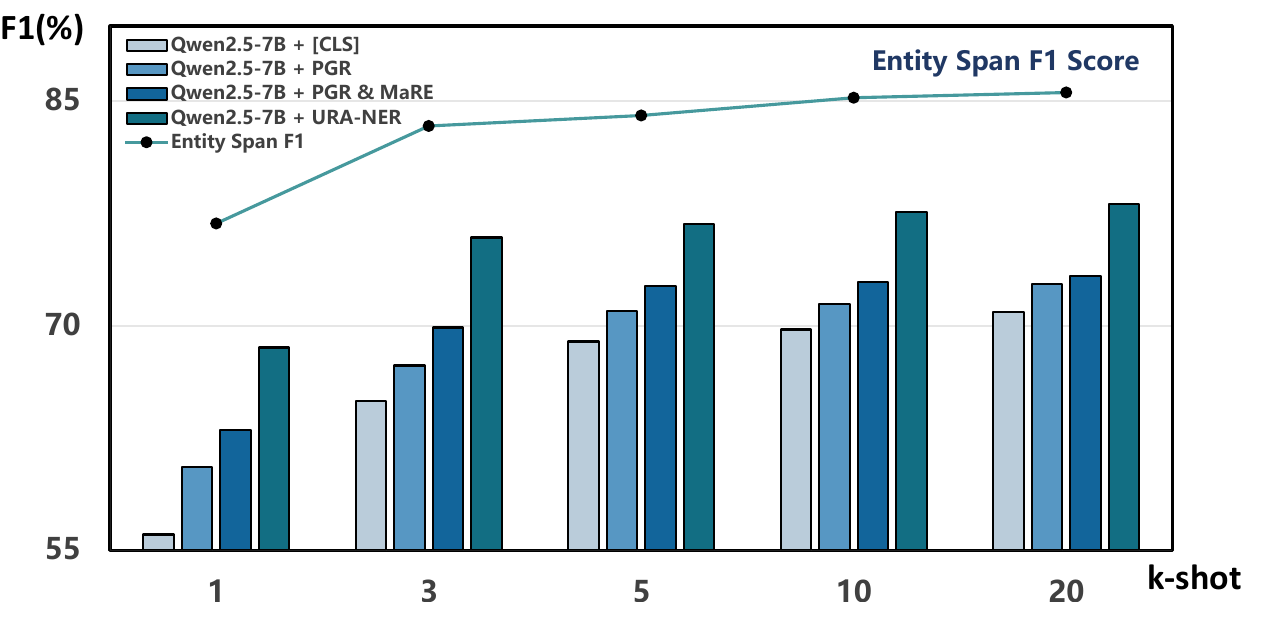}
    \caption{Effect of demonstration numbers in Science.}
    \label{fig:3}
\end{figure}

\subsubsection{Effect of Demonstration Numbers}
Figure~\ref{fig:3} presents the F1 of Qwen2.5-7B under different $k$-shot settings. We compare three configurations: \texttt{+[CLS]}, \texttt{+PGR} and \texttt{+PGR\&MaRE}. We observe that as the number of demonstrations decreases, the performance gap among them widens significantly. This suggests that in extremely low-resource scenarios, stage alignment and representation alignment can greatly improve generalization. Moreover, after introducing RaKV (\texttt{+URA-NER}), the F1 score consistently improves across all $k$-shot settings. Notably, the F1 score in the 1-shot setting is significantly lower than in 3-shot and above, mainly due to suboptimal span extraction in the 1-shot setting. From 3-shot onward, the performance stabilizes. 
This demonstrates that RaKV enhances type classification by retrieving relevant external entity knowledge based on LLM-generated reasoning paths, thereby overcoming the inherent knowledge limitations of the LLM.

 \subsubsection{The Performance of Deepseek-R1}
 Table~\ref{tab:5} reports the 1-shot performance of URA-NER with Deepseek-R1. As the API restricts access to attention weights and hidden states, the related components are disabled ($\lambda=0$) during inference. Compared to \texttt{+[CLS]}, we observe that the performance gain of Deepseek-R1 after incorporating stage alignment (PGR) is relatively small across all six datasets. This is because PGR primarily enhances the type classification capability of LLMs, while Deepseek-R1, as a large-scale model, already exhibits strong performance in this aspect. Therefore, the benefit it gains from PGR is limited. In contrast, for smaller LLMs with clearly weaker type classification ability, PGR proves to be more effective.
Moreover, after incorporating RaKV, Deepseek-R1 exhibits limited performance gains compared to smaller LLMs. This suggests that for extremely large-scale LLMs that possess rich prior knowledge and strong reasoning capabilities, the marginal utility of introducing external knowledge remains limited.

\begin{table}
\centering
\caption{Effect of the order of knowledge chunks.}
\begin{tabular}{lccc}
\toprule
Domain & Reasoning Path & Random & $\Delta$F1 \\
\midrule
Science & 77.76 $\pm$ 0.4 & 76.01 $\pm$ 0.9 & $-$1.75 \\
Politics & 80.40 $\pm$ 0.3 & 77.96 $\pm$ 1.1 & $-$2.44 \\
Music & 83.43 $\pm$ 0.4 & 82.32 $\pm$ 0.7 & $-$1.11 \\
Literature & 74.56 $\pm$ 0.4 & 72.73 $\pm$ 0.9 & $-$1.83 \\
AI & 72.82 $\pm$ 0.3 & 69.84 $\pm$ 1.3 & $-$2.98 \\
CoNLL2003 & 85.59 $\pm$ 0.2 & 85.12 $\pm$ 0.4 & $-$0.47 \\
\bottomrule
\end{tabular}
\label{tab:4}
\end{table}

\subsubsection{The Order of Knowledge Chunks}
 To validate the necessity of reorganizing external knowledge chunks based on the reasoning path, we report in Table~\ref{tab:4} the experimental results of randomly shuffling the retrieved knowledge chunks (\texttt{Random}). Under the 5-shot setting, we compare the average F1 scores of Qwen2.5-7B across six datasets. The results show that randomly disrupting the order of knowledge chunks leads to performance degradation across all datasets, along with a significant increase in standard deviation. This demonstrates that organizing knowledge chunks according to the reasoning path facilitates more effective integration of external information and enhances the stability of model outputs. 

 \subsubsection{Deployment Analysis}

 \begin{table}
\centering
\caption{Deployment analysis of different methods.}
\begin{tabular}{lcccc}
\toprule
Method & Inference & Scale & API cost & F1\\
\midrule
RENER (Gemini-1.5-pro) &  3.9s & $>$100B & $\checkmark$ & 78.3\\
Deepseek-R1 + [CLS] &  9.2s & $>$100B  & $\checkmark$  & 79.9 \\
Qwen2.5-7B + URA-NER & 4.2s & 7B & $\times$  & 79.1\\
Qwen2.5-14B + URA-NER & 5.1s & 14B & $\times$ & 81.4\\
\bottomrule
\end{tabular}
\label{tab:6}
\end{table}

Table~\ref{tab:6} presents the per-query inference time, parameter scale, API cost, and average F1 scores across six datasets for Qwen2.5-7B and Qwen2.5-14B, alongside comparisons with two strong baselines: \texttt{RENER (Gemini-1.5-pro)} and \texttt{Deepseek-R1 + [CLS]}. All experiments are conducted under the 5-shot setting. 
Both URA-NER models (Qwen2.5-7B/14B) achieve competitive F1 scores (79.1\% and 81.4\%) while operating at significantly smaller scales (7B and 14B parameters). Notably, these models eliminate API costs, making them more economical for deployment. In contrast, the Gemini-1.5-pro and Deepseek-R1, which rely on larger than 100B parameters LLMs and incur API expenses. Specifically, \texttt{Qwen2.5-7B + URA-NER} demonstrates a favorable balance between efficiency and performance, with an inference time of 4.2 seconds and an F1 score comparable to the baselines. This analysis highlights the practical benefits of URA-NER in reducing computational and monetary burden while maintaining strong performance and inference efficiency in low-resource NER.

\section{Conclusion}
In this paper, we focus on improving the ICL retrieval mechanism to enhance LLMs' performance in low-resource NER. We propose URA-NER, a novel unified retrieval-augmented framework comprising three key components: PGR for stage-aligned retrieval, MaRE for representation-aligned retrieval, and RaKV to reduce LLM generation uncertainty. Extensive experiments across multiple datasets demonstrate that URA-NER significantly improves performance in low-resource NER, where smaller LLMs outperform existing methods using GPT and other large-scale LLMs, while achieving new state-of-the-art results on CrossNER.






\end{document}